\documentclass{article}

\usepackage{microtype}
\usepackage{graphicx}
\usepackage{subfigure}
\usepackage{caption}
\usepackage{booktabs} 

\usepackage{amsmath}
\usepackage{amssymb}
\usepackage{mathtools}
\usepackage{pifont}

\newcommand{\cmark}{\ding{51}}%
\newcommand{\xmark}{\ding{55}}%

\usepackage{hyperref}

\usepackage[accepted]{icml2019}

\icmltitlerunning{Never Forget: Balancing Exploration and Exploitation via Learning Optical Flow}

\begin{document}

\twocolumn[
\icmltitle{Never Forget: Balancing Exploration and Exploitation \\via Learning Optical Flow}



\icmlsetsymbol{equal}{*}

\begin{icmlauthorlist}
\icmlauthor{Hsuan-Kung Yang}{equal,nthu}
\icmlauthor{Po-Han Chiang}{equal,nthu}
\icmlauthor{Kuan-Wei Ho}{nthu}
\icmlauthor{Min-Fong Hong}{nthu}
\icmlauthor{Chun-Yi Lee}{nthu}
\end{icmlauthorlist}

\icmlaffiliation{nthu}{Elsa Lab, Department of Computer Science, National Tsing Hua University}

\icmlcorrespondingauthor{Hsuan-Kung Yang}{hellochick@gapp.nthu.edu.tw}
\icmlcorrespondingauthor{Chun-Yi Lee}{cylee@cs.nthu.edu.tw}

\icmlkeywords{Machine Learning, ICML}

\vskip 0.3in
]



\printAffiliationsAndNotice{\icmlEqualContribution} 

\begin{abstract}
Exploration bonus derived from the novelty of the states in an environment has become a popular approach to motivate exploration for deep reinforcement learning agents in the past few years.  Recent methods such as curiosity-driven exploration usually estimate the novelty of new observations by the prediction errors of their system dynamics models.  Due to the capacity limitation of the models and difficulty of performing next-frame prediction, however, these methods typically fail to balance between exploration and exploitation in high-dimensional observation tasks, resulting in the agents forgetting the visited paths and exploring those states repeatedly.  Such inefficient exploration behavior causes significant performance drops, especially in large environments with sparse reward signals.  In this paper, we propose to introduce the concept of optical flow estimation from the field of computer vision to deal with the above issue.  We propose to employ optical flow estimation errors to examine the novelty of new observations, such that agents are able to memorize and understand the visited states in a more comprehensive fashion.  We compare our method against the previous approaches in a number of experimental experiments.  Our results indicate that the proposed method appears to deliver superior and long-lasting performance than the previous methods.  We further provide a set of comprehensive ablative analysis of the proposed method, and investigate the impact of optical flow estimation on the learning curves of the DRL agents.

\end{abstract}

\begin{figure}
    \hskip -0.3in
    \includegraphics[width=1.1\columnwidth]{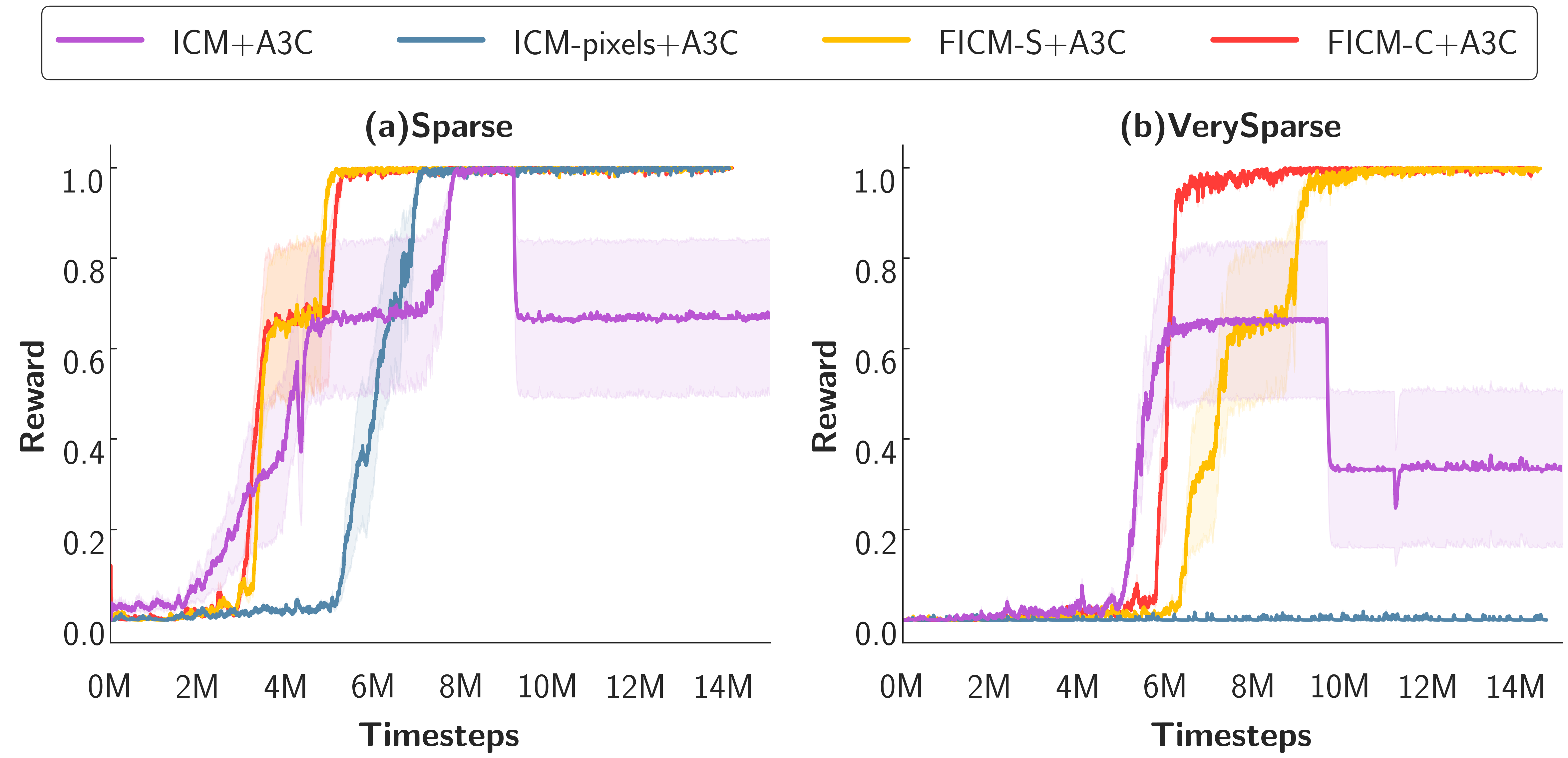}
    \caption{
    Comparison of the learning curves of two ICM baselines and the proposed methods on ViZDoom. The learning curve of the purple line clearly shows that catastrophic forgetting impairs the performance of `ICM + A3C' after a long period of timesteps during the training phase. In this paper, we introduce flow-based ICM (FICM) for measuring the novelty of states in exploration.  The yellow and red lines show that the two implementations of FICM enable better and stabler performances than those of ICM.}
    
    \label{fig:reward}
    \vskip -0.1in
\end{figure}

\section{Introduction}
\label{submission}

Reinforcement learning (RL) algorithms are aimed at developing policies to maximize rewards provided by the environment, and retrieve major success in variety of application domains, such as game playing~\cite{mnih2015human,silver2016mastering} and robot navigation~\cite{zhang2016learning}. The usual RL methods require operations under a shaped reward function, but in real world situation there is few environments returning satisfying shaped reward all the time. In some scenarios, rewards extrinsic to the agent are extremely sparse, and it is impossible to engineer dense reward functions under the circumstances. Adopting simple heuristic methods is a common solution, such as $\epsilon$-greedy~\cite{mnih2015human,sutton1998reinforcement} or entropy regularization~\cite{mnih2016a3c}. However, strategies like such are still far from generating satisfactory results in tasks with deceptive or sparse rewards~\cite{fortunato2018noisy,osband2017deep}. 

Researchers in recent years come up with solutions to those environments with sparse rewards by providing agents with bonus rewards whenever they visit an unexplored or unexpected observation. These bonus rewards tend to encourage the agent explore the states that are not seen before, i.e. novel state, and measuring the ''novelty'' usually requires a additional model to measure the statistic distribution of the environmental states. For example,~\cite{houthooft2016vime} uses information gain as a measurement of state novelty. In~\cite{tang2017exploration}, a counting table is used to estimate the novelty of a visited state. Neural density model~\cite{van2016conditional} is also served to measure bonus rewards for agents. In~\cite{stadie2015incentivizing,ICM}, the novelty of a state is estimated from the prediction errors of their system dynamics models. In the proposed curiosity-driven exploration~\cite{ICM}, curiosity helps an agent explore its environment in the quest for new knowledge. The curiosity is formulated as the error in an agent’s ability to predict the consequence of its own actions in a visual feature space, and this process is learned by intrinsic curiosity module (ICM). In proposed exploration bonus \cite{yuri2018exploration}, an exploration bonus is the error of a neural network predicting features of the observations given by a fixed randomly initialized neural network. The random network distillation (RND) bonus combined with a method to flexibly combine intrinsic and extrinsic rewards enables significant progress on several hard exploration Atari games.

However, as we deep-dive into curiosity-based exploration~\cite{ICM} in ViZDoom experiment, the catastrophic forgetting problem~\cite{french1999catastrophic} occurs. In Fig.~\ref{fig:reward}, we can see clearly that after 10 million timesteps, the curiosity-based agent employing ICM suffers from serious performance drop on two experiments with different sparsity. This performance drop problems is often known as catastrophic forgetting~\cite{mccloskey1989catastrophic,robins1995catastrophic,sharkey1995analysis,french1999catastrophic} and is common in machine learning. The occurrence of catastrophic forgetting is often related to the fact that the model requires sufficient plasticity to digest new information, but large weight changes cause forgetting by disrupting previous learned representations. Such performance drop shows the sign of unstable performance of the curiosity-based exploration method. Furthermore, we also look into the value changes of curiosity, and we found out the sudden increase of the curiosity after 10 million timesteps. This discovery clearly indicates that after numerous timesteps of exploring and training, the agent finally forgets previous explorations and misunderstands those seen states as novel states. Both discoveries above suggest that the curiosity-based exploration agent suffers from catastrophic forgetting problem and there are room for improvements. Researchers in recent years proposed solutions~\cite{goodrich2014unsupervised,ian2015empirical,ronald2018measuring} to the catastrophic forgetting, and they name these solutions lifelong learning. Most lifelong learning techniques usually include regularization, multi-memory deployment and ensemble methods to ensure the early representation is stably updated and mixed with the later learned representation peacefully. However, to the best of our knowledge, there are few works for directly addressing the catastrophic problem in RL.  

We consider that the catastrophic forgetting problem of curiosity-based exploration is due to its attempt at next frame prediction. With the analysis mentioned above, we try to find out the problem responsible for catastrophic forgetting in curiosity-based exploration design and to improve it. In the curiosity-based exploration, the policy is trained to optimize the sum of the extrinsic reward and intrinsic reward. The extrinsic reward is provided by the environment and the curiosity-based intrinsic reward is generated by the proposed ICM~\cite{ICM}. ICM is composed of two parts, which are inverse dynamics model and forward model. The inverse dynamics model takes encoded features of current state and next state as inputs, so as to predict the action to take from current state to desired next state; the forward model takes encoded current state and the action as feature to predict the next state feature representation. The discrepancy between predicted next state feature and actual next state encoded feature is the curiosity-based intrinsic reward. We reckon that the feature of the forward model's next frame prediction, which is solely based on current state and action, is too simple to be reliable. In computer vision or other machine learning field, next frame prediction requires deep neural networks or complex features~\cite{vae,gan,cgan,deep-pred-coding-net,visual-dynamics} to output an entire predicted scene features, and the features used in forward model are clearly not that complex. The simplicity of forward model is possibly the reason the agent suffers from serious forgetting problem after long time of learning.

In order to solve the catastrophic forgetting problem, we propose solutions from different angles by introducing optical flow~\cite{FlowNet2,Unflow}, which is a popular techniques in computer vision field to comprehend continuous movements of objects and environments, into the exploration design. The optical flow is a powerful techniques for computer vision researcher to extract features from and analyze continual materials, such as videos or continual frames of images. The optical flow extracts movements and discrepancy between frames in a way more complex and meaningful than methods such as next frame prediction or straightforward frame difference that other proposed exploration bonus methodologies tend to adopt. In this work, we replace ICM in the original curiosity-based exploration design~\cite{ICM} with our own version of intrinsic reward generator --- flow-based intrinsic curiosity module (FICM). Our proposed design with optical flow suffers no sudden drop of performance, consumes fewer parameter, and reaches better and lasting performance in several experiments. Moreover, we even present two different architectures of the flow predictor inside FICM to verify the effectiveness of optical flow.


     In order the demonstrate the effectiveness of FICM, we perform experiments on the ViZDoom~\cite{ViZDoom} environment with sparse reward and very sparse reward settings.  We show that the propose FICM outperforms the previous ICM with better and stabler performance even after a long period of training timesteps, and does not suffer from the catastrophic forgetting problem which occurs in ICM for both settings.  We also show that FICM enables an RL agent to learn faster, and requires fewer number of model parameters (180K $\sim$ 250K) than ICM.  We further illustrate the evolution of intrinsic rewards versus time to demonstrate the robustness of FICM to catastrophic forgetting.  We provide a comprehensive analysis of our experimental results and investigate the impacts of FICM in Section~\ref{sec::experimental_results}.  The primary contributions of this paper are summarized as the following:


\begin{itemize}
\itemsep=-2pt

\item We identify and validate the existence of catastrophic forgetting in curiosity-based exploration, which causes sudden drops in performance after a long periods of training timesteps. 
\item We propose an enhanced version of ICM, called FICM, to deal with the catastrophic forgetting problem of ICM in RL exploration.  FICM is based on optical flow estimation, a popular technique in computer vision to estimate movements of objects in consecutive frames.
\item The proposed FICM only requires state information, and does not need actions of the agents when estimating the novelty of states.
\item The proposed FICM requires significantly fewer number of model parameters than ICM, and is still able to deliver superior performance to it.
\item The proposed FICM takes only two consecutive image frames as its input, which is significantly fewer than that of ICM.
\end{itemize}

This rest of this paper is organized as follows. Section~\ref{sec::background} introduces background material. Section~\ref{sec::methodology} walks through the proposed framework and its implementation details.  Section~\ref{sec::experimental_results} presents the experimental results and provides a comprehensive analysis of the framework. Section~\ref{sec::conclusion} concludes.

\begin{table}[t]
\centering
\caption{Notation used in this paper}
\vskip 0.2in

\footnotesize
\begin{tabular}{l|l}
\hline
$t$ & Timestep index                     \\
\hline
$S_{t}$ & State at timestep $t$  \\
${r^{e}}$ & Extrinsic reward \\
${r^{i}}$ & Intrinsic reward \\
$r^{f}$ & Reward generated from forward flow\\
$r^{b}$ & Reward generated from backward flow\\
$L^{f}$ & Loss from forward flow \\
$L^{b}$ & Loss from backward flow \\
$\Theta_{\pi}$ & Parameters of A3C policy \\
$\Theta_{f}$ & Parameters of flow predictor \\
\hline
\end{tabular}
\label{notation}
\vskip 0.1in
\end{table}

\begin{figure*}[t]
\begin{minipage}{.3\textwidth}
      \centering
      \includegraphics[height=2.0in]{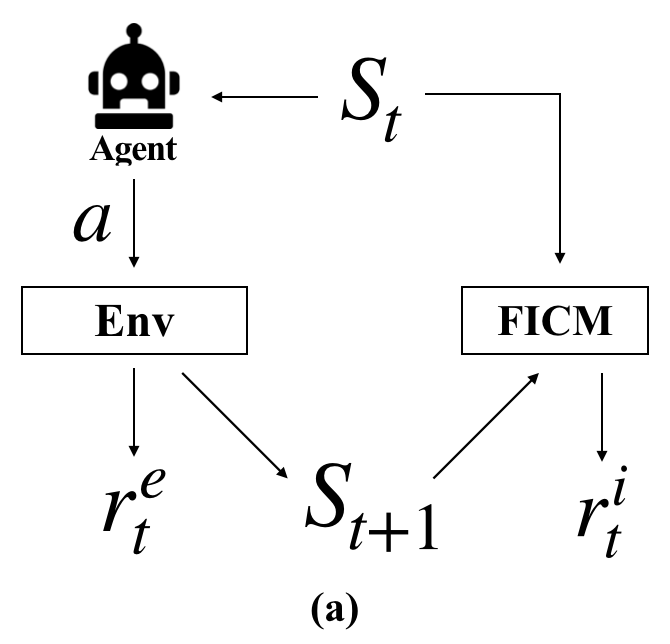}
    \end{minipage}%
    \begin{minipage}{.7\textwidth}
      \centering
      \includegraphics[height=2.1in]{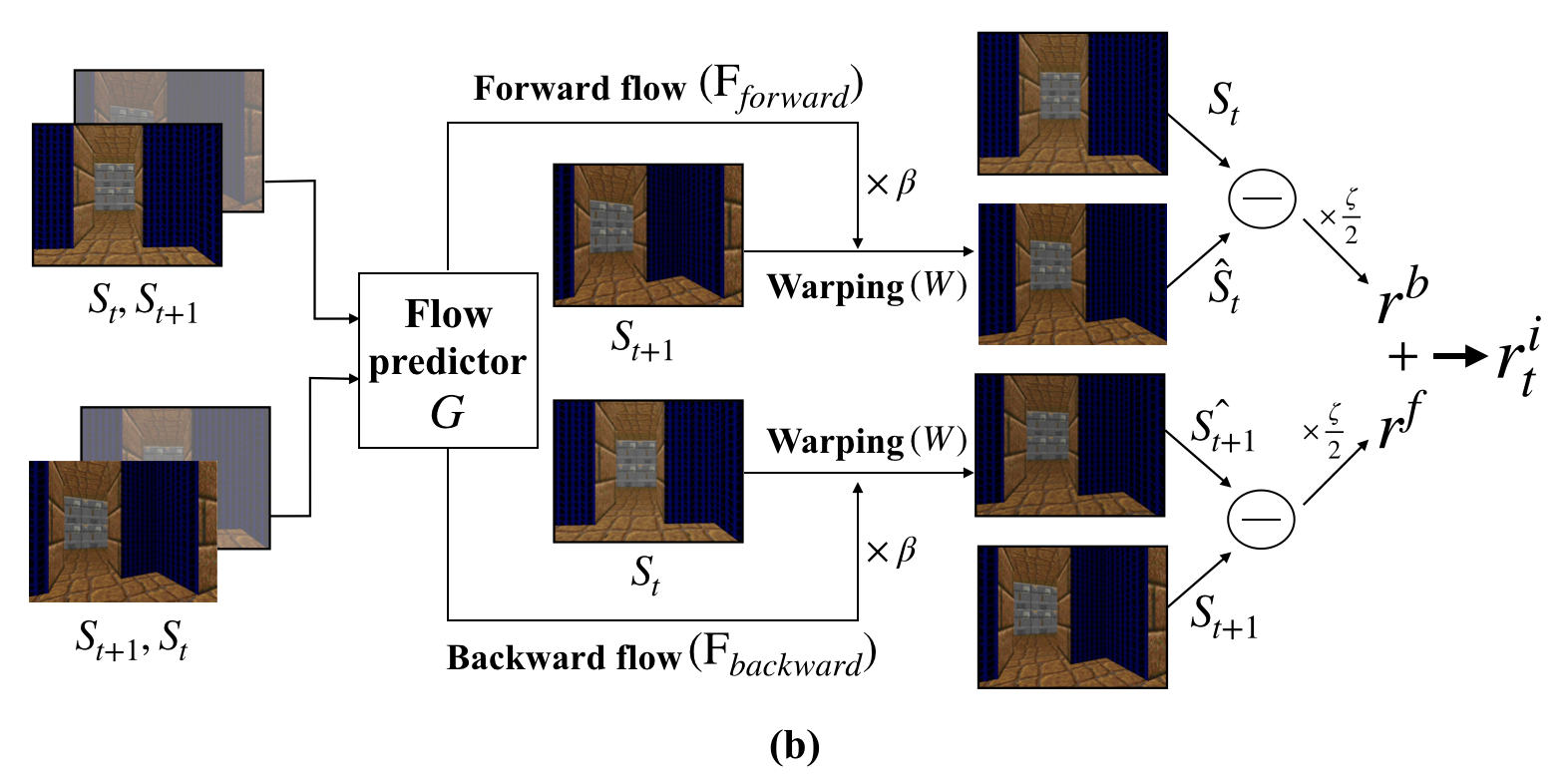}
    \end{minipage}
\vskip -0.1in
\caption{Framework overview. (a) Illustration of the interactions among FICM, the agent, and the environment.  (b) The workflow of FICM.
}
\label{fig:overviews}
\end{figure*}

\section{Background}
\label{sec::background}
In this section, we introduce background material. We
first provide an overview of deep reinforcement learning (DRL) and curiosity-driven exploration. To understand the proposed FICM, we also briefly review related works that focus on optical flow estimation.

\vskip 0.2in
\subsection{Deep Reinforcement Learning}
DRL is a method to train an agent to interact with an environment $\mathcal{E}$. 
At each timesteps t, the agent recieves a state \textit{$s_t$} form the state space $\mathcal{S}$ of $\mathcal{E}$, takes an action $a_t$ from the action space $\mathcal{A}$ according to its policy $\pi(a|s)$, receives a reward $r(s_t, a_t)$, and transits to next state $s_{t+1} \sim p(s_{t+1}|s_t, a_t)$.
The main objective of agent is to maximize discounted cumulative rewards $R_t = \sum_{i=t}^{T} \gamma^{i-t}r(s_i, a_i)$, where $\gamma \in (0, 1]$ is a discont factor and T is the horizon. The action-value function (i.e., Q-function) of a given policy $\pi$ is defined as the expected return starting from a state-action pair $(s, a)$, expressed as $Q(s, a) = \mathbb{E}[R_t|s_t = s, a_t = a, \pi]$. With the flourish development in Deep Neural Network (DNN), Deep Q-learning (DQN) ~\cite{DQN} takes advantages of DNN to represent the Q-function. In addition, Asynchronous Advantage Actor-Critic (A3C) ~\cite{A3C} further introduces asynchronous calculation and optimization to update its policy and value function.


\subsection{Curiosity-Based Exploration}
Curiosity-based exploration is an exploration strategy adopted by some DRL agents in recent works \cite{houthooft2016vime, ICM, yuri2018large} in order to explore environment more efficiently. The traditional random exploration often has hard time being trapped in local minima of state spaces while curiosity-based tends to seek out relatively unexplored regions, and therefore is able to explore more productively in the same amount of time. In addition to extrinsic reward provided by environment, most curiosity-based exploration works introduce intrinsic reward generated by the agent to encourage itself to explore novel states. For example in the proposed ICM \cite{ICM}, it takes frame prediction error as intrinsic reward, which suggests that the agent will receives more reward if next state is beyond its knowledge about the environment, and thus is more `curious' to explore novel states.

\subsection{Optical Flow Estimation}
Optical flow estimation \cite{FlowNet} is a technique to evaluate the motion of objects between between consecutive images. In usual cases, a reference image and a target image are required. The optical flow is represented as a vector field, where displacement vectors are assigned to certain pixels of the reference image. These vectors represent where those pixels can be found in the target image.  In recent years, a number of deep learning approaches running on GPUs dealing with large displacement issues of optical flow estimation have been proposed \cite{FlowNet, DeepFlow, FlowNet2}.  Among these techniques, FlowNet 2.0 \cite{FlowNet2} delivers the most accurate estimation. In this paper, we use simplified network of FlowNet 2.0 to generate optical flow.

\section{Methodology}
\label{sec::methodology}
In this section, we present the design and implementation details of our methodology. We first provide an overview of the proposed framework, followed by an introduction to the fundamental concepts of FICM.  Then, we formally formulate these concepts into mathematical equations, and walk through the details of our training objective.  Finally, we investigate two different implementations of FICM, and discuss the features and advantages of their configurations.


\subsection{Framework Overview}
\label{subsec::framework_overview}
Fig.~\ref{fig:overviews} illustrates an overview of the proposed methodology. In our framework, the agent moves according to a policy which is trained to perform actions to maximize the total rewards it receives. The rewards that the agents receives comes from either the extrinsic reward $r^e$ or the intrinsic reward $r^i$.  The extrinsic reward $r^e$ is obtained from the agent's interaction with the environment.  On the other hand, the intrinsic reward $r^i$ is generated by FICM to encourage the agent to explore the environment.
FICM is a deep neural networks composed of a number of convolutional and deconvolutional layers, and designed to utilize optical flow to estimate the novelty of states. Based on the framework, FICM gradually learns the explored state more comprehensively over time. FICM yields a higher reward when visiting a state considered to be more novel, and generates more incentives for the agent to pursue novel observations. Details of novelty measurement and further descriptions are provided in Sections~\ref{subsec::flow_based_curiosity_driven} and~\ref{subsec::FICM}. The policy parameterized by $\Theta_{\pi}$ is trained using policy learning methods to maximize the total rewards.  In this paper, A3C is adopted as our policy learning method.

\subsection{Flow-based Curiosity Driven Exploration}
\label{subsec::flow_based_curiosity_driven}


We consider that the primary reason of the catastrophic forgetting problem in ICM lies in the difficulty to perform next frame prediction. In ICM, the novelty of states are evaluated based on the prediction error of the forward dynamics model.  As predicting the next state or its features simply from the current state and action is not straightforward, the forward dynamics model tend to fail in large and high-dimensional state spaces due to insufficient information to make such predictions.  As a result, the forward dynamics model in ICM tends to "forget" the states that the RL agent has visited before, leading to a severe performance drop when the number of training steps increases.


We propose to employ optical flow estimation, which is a popular techniques in the field of computer vision to comprehend continuous movements of objects and environments, to address the catastrophic forgetting problem. Optical flow estimation is a powerful technique for computer vision researchers to extract features from and analyze continuous frames of images. Similar to the inverse dynamics model in ICM, optical flow also captures delicate changes between frames, which makes itself a suitable option for estimating the novelty of states. In the inverse dynamic model design, the model generates a high intrinsic reward when it encounters an observation that is far different from its prediction based on the current state and action, and a low intrinsic reward is generated when the encountered observation is expected. Similarly, FICM yields high intrinsic rewards when it observes unfamiliar flow changes, 
and low intrinsic rewards when the patterns of optical flow are learned before. If FICM fails to predict the optical flow between the current state and the next state correctly, the next state is then considered as a novel state.  FICM then generates a intrinsic reward signal to motivate the RL agent to explore that state.

In Fig.~\ref{fig:overviews}, the left part shows FICM's role in the interaction between agent and environment described above, and the right part shows the entire workflow of FICM. FICM takes two consecutive states as its input, and predicts a forward flow $F_{forward}$ and a backward flow $F_{backward}$ from the pairs of input states. The forward flow $F_{forward}$ is the optical flow inferenced from consecutive states ordered in time increasing fashion (e.g., $t$ to $t+1$), while the backward flow is the optical flow inferenced from consecutive states ordered in time decreasing fashion  (e.g., $t+1$ to $t$). The input states $S_t$ and $S_{t+1}$ are then warped by the flows to generate the predicted states $\hat{S_t}$ and $\hat{S_{t+1}}$.  The losses of these predicted states serve as the partial intrinsic reward signals $r^b$ and $r^f$, respectively.  The sum of $r^f$ and $r^b$ forms the final intrinsic curiosity reward $r^i$ outputted by FICM. More design and implementation details of FICM are provided in Section~\ref{subsec::FICM}.


\subsection{Flow-based Intrinsic Curiosity Module}
\label{subsec::FICM}

In this section, we formulate the procedure of FICM as formal mathematical equations.  The main objective of FICM is to leverage the optical flow between two consecutive states as the encoded representation of them.  As described in Sections~\ref{subsec::framework_overview} and~\ref{subsec::flow_based_curiosity_driven}, FICM predicts a forward flow $F_{forward}$ and backward flow $F_{backward}$ from each consecutive pairs of input states. Given two raw input states $S_{t}$ and $S_{t+1}$ observed at consecutive timesteps $t$ and $t+1$, FICM takes the 2-tuple $(S_{t}, S_{t+1})$ as its input, and predicts $F_{forward}$ and $F_{backward}$  by its flow predictor $G$ parameterized by a set of trainable parameters $\Theta_f$.  The two flows $F_{forward}$ and $F_{backward}$ can therefore be expressed as the following:



\begin{equation}
\begin{aligned}
&\mathrm{F_{forward}} = G(\mathrm{S_{t}}, \mathrm{S_{t+1}}, \Theta_f) \\
&\mathrm{F_{backward}} = G(\mathrm{S_{t+1}}, \mathrm{S_{t}}, \Theta_f).
\label{equation1}
\end{aligned}
\end{equation}


$F_{forward}$ and $F_{backward}$ are then used to generate the predicted states $\hat{S_t}$ and $\hat{S_{t+1}}$ via a warping function $W$~\cite{FlowNet2}. The predicted $\hat{S_t}$ and $\hat{S_{t+1}}$ are expressed as:

\begin{equation}
\begin{aligned}
&\mathrm{\hat{S_{t}}} = W(\mathrm{S_{t+1}}, \mathrm{F_{forward}}, \beta) \\
&\mathrm{\hat{S_{t+1}}} = W(\mathrm{S_{t}}, \mathrm{F_{backward}}, \beta),
\label{eq::warping}
\end{aligned}
\end{equation}


where $\beta$ is the flow scaling factor.  $W$ warps $\mathrm{S_{t+1}}$ to ${\hat{S_{t}}}$ and $\mathrm{S_{t}}$ to ${\hat{S_{t+1}}}$ via $\mathrm{F_{forward}}$ and $\mathrm{F_{backward}}$ respectively using bilinear interpolation and element-wise multiplication with $\beta$.  The interested reader is referred to~\cite{FlowNet, FlowNet2} for more details of the warping algorithm.  Please note that in this work, $W$ employs inverse mapping instead of forward mapping to avoid the common duplication problem in flow warping~\cite{thaddeus1992feature}.


With the predicted $\mathrm{\hat{S_{t}}}$ and $\mathrm{\hat{S_{t+1}}}$, $\Theta_f$ is iteratively updated to minimize the loss function $L_G$ of $G$, which consists of a forward loss $L^f$ and a backward loss $L^b$. $L_G$ is written as:


\begin{equation}
\begin{aligned}
\min_{\Theta_f}L_G &= L^f + L^b \\
&= {||\mathrm{S_{t+1}} - \mathrm{\hat{S_{t+1}}}||}^2 + {||\mathrm{S_{t}} - \mathrm{\hat{S_{t}}}||}^2,
\end{aligned}
\label{eq::loss_function}
\end{equation}

where $(L^f, L^b)$ are derived from the mean-squared error (MSE) between $(\mathrm{S_{t+1}}, \hat{S_{t+1}})$ and $(\mathrm{S_{t}}, \hat{S_{t}})$, respectively.  In this work, $L_G$ is interpreted by FICM as a measure of state novelty, and serves as an intrinsic reward signal $r^i$ for the DRL agent in Fig.~\ref{fig:overviews}.  The expression of $r^i$ is represented as:

\begin{equation}
\begin{aligned}
r^i &= r^f + r^b = \frac{\zeta}{2}(L^f + L^b) = \frac{\zeta}{2}L_G \\
&=\frac{\zeta}{2}({||\mathrm{S_{t+1}} - \mathrm{\hat{S_{t+1}}}||}^2 + {||\mathrm{S_{t}} - \mathrm{\hat{S_{t}}}||}^2),
\end{aligned}
\label{eq::intrinsic_reward}
\end{equation}

where $\zeta$ is the reward scaling factor, and $r^f$ and $r^b$ are the forward and backward intrinsic rewards scaled from $L^f$ and $L^b$, respectively.  Please note that $r^i$ is independent of the action taken by the agent, which distinguishes FICM from ICM.  FICM only takes two consecutive raw input states for estimating the optical flow, which serves as a more meaningful measure to evaluate and memorize the novelty of states in large high-dimensional state spaces with sparse external reward signals.  The experimental results presented in Section~\ref{sec::experimental_results}
demonstrate the effectiveness of $r^i$ and FICM.


\begin{figure}[t]
\begin{center}
\centerline{\includegraphics[width=\linewidth]{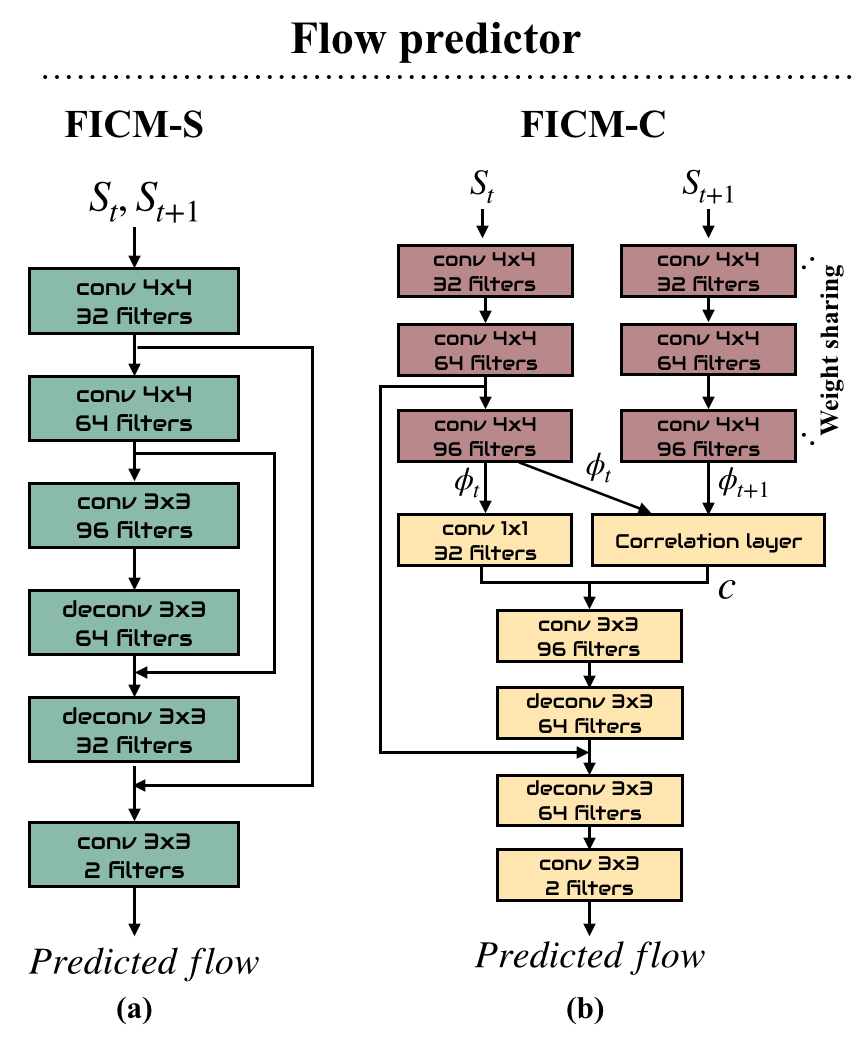}}
\caption{The flow predictor architectures in FICM-S and FICM-C.}
\label{fig:flow_archi}
\end{center}
\vskip -0.3in
\end{figure}


\subsection{Implementations of FICM}
\label{subsec::implementation_FICM}
In this work, we propose two different implementations of FICM: FICM-S and FICM-C.  These two implementations adopt different flow predictor architectures based on FlowNetS and FlowNetC employed by FlowNet~2.0~\cite{FlowNet2}, respectively. Different implementations are introduced to validate that $r^i$ derived from $L_G$ based on Eq.~(\ref{eq::intrinsic_reward}) can indeed serve as a suitable intrinsic reward signal, rather than restricted to FICMs with specific flow predictor architectures.
The flow predictor architectures are depicted in Fig.~\ref{fig:flow_archi}, and explained in the following paragraphs.



\textbf{FICM-S.}~~ 
The flow predictor in FICM-S consists of several convolutional and deconvolutional layers. The module first stacks two consecutive states $S_t$ and $S_{t+1}$ together, and feed the stacked states $\left\langle S_{t}, S_{t+1}\right\rangle$ into three convolution layers with 32, 64, and 96 filters, respectively, followed by an exponential linear unit (ELU) non-linear activation function.  The encoded features are then fused with the feature maps from the shallower parts of the network by adding skips~\cite{FlowNet, FCN}, and fed into two deconvolutional layers with 64 and 32 filters.  This skip layer fusion architecture allows the flow predictor to preserve both coarse, high layer information and fine, low layer information~\cite{FlowNet, FCN}.  Finally, the feature map is passed into a convolutional layer with two filters to predict the optical flow from $S_t$ to $S_{t+1}$.


\textbf{FICM-C.}~~
The flow predictor in FICM-C encodes two consecutive states $S_t$ and $S_{t+1}$ separately instead of stacking them together. The input states are passed through three convolutional layers to generate feature maps $\phi_{t}$ and $\phi_{t+1}$.  The three convolutional layers of the two paths are share-weighted in order to generate better representations of $\phi_{t}$ and $\phi_{t+1}$, as input states $S_t$ and $S_{t+1}$ usually contain same or similar patterns. The feature maps $\phi_{t}$ and $\phi_{t+1}$ are then fed into a correlation layer proposed by~\citet{FlowNet}, which performs multiplicative patch comparisons between two feature maps to estimate their correspondences $c$ defined as:


\begin{equation}
c(x_1,x_2)=\sum_{\mathclap{o\in{[-k,k]\times[-k,k]}}}\left\langle(\phi_{t}(x_1+o), \phi_{t+1}(x_2+o)\right\rangle,
\end{equation}

where $x_1$ is the patch center in the first feature map and $x_2$ is that in the second one. Here we fix the maximum displacement $d$ to $2$, which means we only compute correlations $c(x_1, x_2)$ limited in a neighborhood of size $D := 2d + 1$ by constraining the range of $x_2$. Since the feature maps of our models have moderate resolutions, $d=2$ is sufficient to find their correspondences.

Once the correspondences $c$ of $\phi_{t}$ and  $\phi_{t+1}$ is estimated, it is concatenated with the feature map from the fourth convolutional layer of the left path in Fig.~\ref{fig:flow_archi}(b). The concatenated feature map is later passed through one convolutional and one deconvolutional layers, fused with the feature map came from the skip path, and then passed through another deconvolutional layer. Similar to Fig.~\ref{fig:flow_archi}(a), the final feature map in Fig.~\ref{fig:flow_archi}(b) is fed into a convolutional layer with two filters to predict the optical flow from $S_t$ to $S_{t+1}$.


After estimating the optical flow from the flow predictor of either FICM-S or FICM-C, the flow is used to warp the states $S_t$ and $S_{t+1}$ forward and backward using Eq.~(\ref{eq::warping}), respectively, and derive $L_G$ and $r^i$ according to Eqs.~(\ref{eq::loss_function}) and~(\ref{eq::intrinsic_reward}).


\begin{figure}[t]
    \centering
    \includegraphics[width=\linewidth]{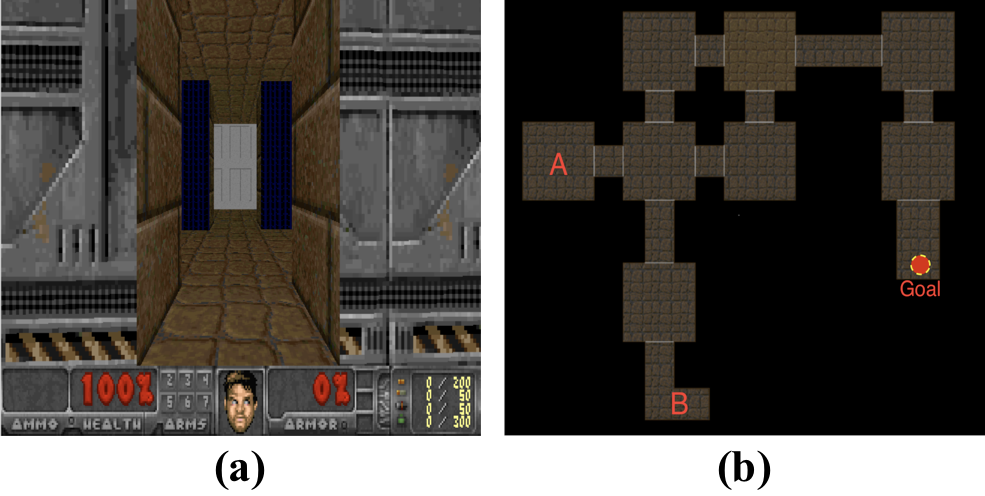}
    \caption{(a) The raw input image from ViZDoom. (b) The map of our environment. Please note that the points A and B are the initial spawning locations in the "sparse reward" and "very sparse reward" settings, respectively.}
    \label{fig:Raw_input}
\end{figure}

\section{Experimental Results}
\label{sec::experimental_results}
In this section, we present experimental results and discuss their implications.  We start by a brief introduction to our experimental setup, including the environment, the baselines, and their training methodologies.  Next, we compare the proposed methodology against the previous approaches in qualitative and quantitative experiments.  We prove that FICM outperforms the previous ICM in two aspects: (1) better and stabler performance of the learned policy evaluated on ViZDoom~\cite{ViZDoom}; and (2) no suffering from the catastrophic forgetting problem occurring in ICM.




\subsection{Environmental Setups}
\textbf{Environment.}~~The environment we evaluate on is the ViZDoom \cite{ViZDoom} game. ViZDoom is a popular Doom-based platform for AI research especially reinforcement learning to test agents' ability to process raw visual information shown in Fig.~\ref{fig:Raw_input}(a). We conduct experiments on the same gaming environment, \textit{DoomMyWayHome-v0}, as in the proposed ICM~\cite{ICM}. The game map consists of 9 rooms and the agent is tasked to reach some fixed goal locations from its spawning location. The agent is only provided with a sparse terminal reward of +1 if it finds the vest and zero otherwise. The agent consists of four discrete actions - move forwards, move left, move right and no action. In order to evaluate the exploration ability under sparse extrinsic reward environment, we adopt two setups which are also experimented in ICM --- `sparse' and `very sparse' rewards. As shown in Fig.~\ref{fig:Raw_input}(b), both rewards are defined based on their distances between the initial spawning location of the agent and the location of the goal. The further the goal is from the spawning location, the harder for the agent to explore.

\textbf{Baseline approach}~~
We compare FICM with two baselines, `ICM + A3C' and `ICM-pixels + A3C' which are both proposed in~\cite{ICM}. ICM combines intrinsic curiosity module with A3C. ICM-pixels is close to ICM in architecture except without the inverse dynamics model, and ICM-pixels computes curiosity reward only dependent on forward model loss in next frame prediction.

\textbf{Training methodology}~~
This section describes the implementation details of our training methodology. The agents are trained directly with raw image frames captured from ViZDoom, as shown in Fig.~\ref{fig:Raw_input}. All input frames are converted from RGB into gray-scale and resized to $42 \times 42$.  Please note that ICM requires eight such frames (four stacked frames for $S_t$ and another four for $S_{t+1}$) as its input.  On the other hand, both FICM-S and FICM-C require only a single frame for $S_t$ and another one for $S_{t+1}$.  At each step of interaction with the ViZDoom environment, an action is repeated four times during the training phase.  Following the asynchronous training algorithm described in~\citet{A3C}, the four agents `ICM +A3C', `ICM-pixels + A3C', `FICM-S + A3C' and `FICM-C + A3C' are trained asynchronously in the 'sparse' and 'very sparse' reward settings described above with twenty workers using stochastic gradient descent and ADAM optimizer for 15M timesteps.



\subsection{Comparison of the Learning Curves}
\label{subsec::learningcurve}
We analyze and compare the learning curves of our proposed methods, `FICM-C + A3C' and `FICM-S + A3C', with those of the baseline methods, `ICM + A3C' and `ICM-pixels + A3C'.  Fig.~\ref{fig:reward} shows the learning curves.  Each curve is plotted from three independent runs, and smoothed among timesteps.  The solid lines represent the means of the curves, while the shaded areas represent the confidence intervals. We compare the results in the sparse reward setting and very sparse reward settings in the following paragraphs.

\textbf{Sparse reward setting.}~~
In the experimental results presented in~\citet{ICM}, both `ICM + A3C' and `ICM-pixels + A3C' learn well, and are able to achieve the maximum performance before 9M timesteps. However, according to our experiments, `ICM + A3C' begins suffering from the catastrophic forgetting problem at 10M timesteps, leading to a severe drop in performance.  In contrast, `FICM-C + A3C', `FICM-S + A3C', and `ICM-pixels + A3C' maintain the performance even at 15M timesteps.  It can also be observed that our methods converge faster than the baselines.


\textbf{Very sparse reward setting.}~~
In the very sparse reward setting, `ICM + A3C' continues to suffer from the catastrophic forgetting problem, while `ICM-pixels + A3C' even fails in this setting.  Although both `FICM-C + A3C' and `FICM-S + A3C' do not converge faster than `ICM + A3C', they are able to maintain their performance till the end of the entire training process, without any observable performance drop.


We can observe that no performance drop occurs for `ICM-pixels + A3C' under the sparse reward setting in Fig.~\ref{fig:reward}(a).  One possible explanation is that the overly strong exploration capability offered by ICM may actually end up trading exploitation off for exploration.  In other words, if the agent is eager to explore novel states without balancing exploration and exploitation, it is more likely to forget the previously explored states.  Moreover, as the reward signals in the sparse reward setting is relatively dense than the those in the very sparse setting, `ICM-pixels + A3C' is able to play well even with a weaker exploration capability.  On the contrary, the overly strong exploration capability of `ICM + A3C' causes it to be vulnerable to the catastrophic forgetting problem.  `FICM-C + A3C' and `FICM-S + A3C' are able to balance well between exploration and exploitation, enabling them to be more robust to the forgetting problem.




\begin{table}[t]
\caption{Comparison of the number of parameters and the existence of catastrophic forgetting for different ICM methods.  Please note that `-' indicates that the corresponding method fails the task.}
\label{param_table}
\begin{center}
\begin{footnotesize}
\begin{sc}
\resizebox{\columnwidth}{!}{
\begin{tabular}{lcccc}
\toprule
Method & \# of parameters  & \multicolumn{2}{c}{Forgetting} \\
\midrule
 &  & Sparse & Very sparse \\ 
ICM     & 326,692 &  \cmark & \cmark \\
ICM-pixels  & 142,212 & \xmark & -- \\
FICM-C (ours) & 257,570 & \xmark & \xmark \\
FICM-S (ours) & 182,594 & \xmark & \xmark \\
\bottomrule
\end{tabular}
}
\end{sc}
\end{footnotesize}
\end{center}
\vskip -0.1in
\end{table}

\begin{figure*}[t]
    \vskip 0.2in
    \centering
    \includegraphics[width=\textwidth]{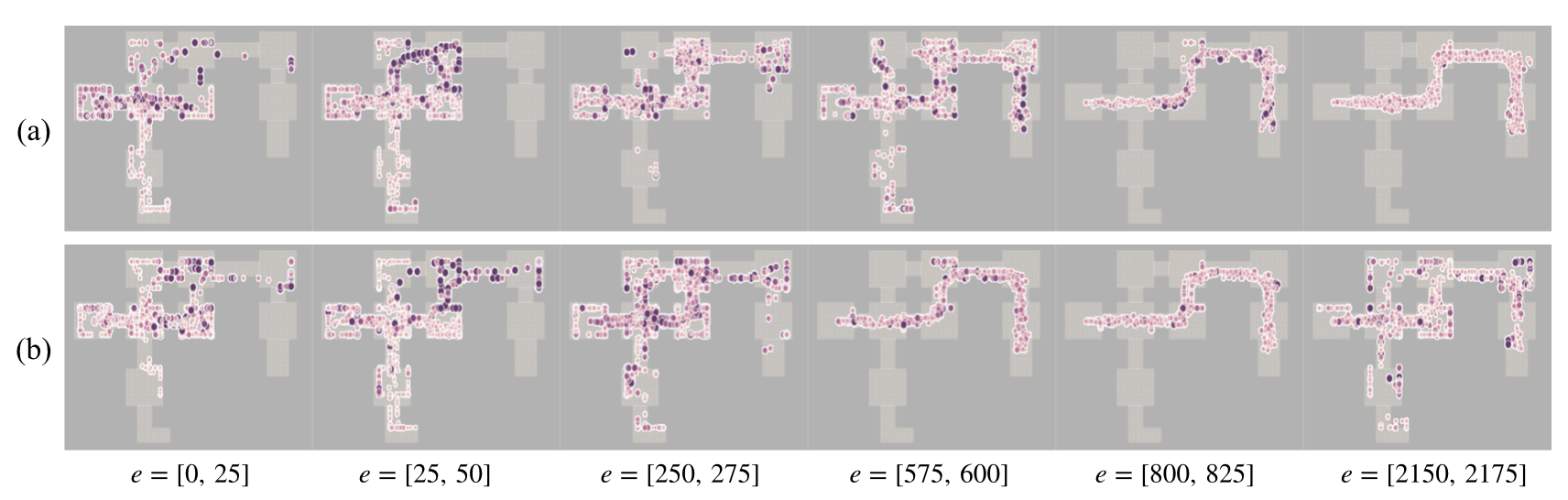}
    \vskip -0.1in
    \caption{Evolution of intrinsic rewards versus time.
    The 3-tuples $(x,y,bonus)$ from different episodes are plotted on the game map.  Episode range $e=[0,25]$ means that the tuples from episodes 0 to 25 are plotted. Darker points correspond to higher exploration bonuses (i.e., intrinsic rewards). (a) The footprints of FICM, and (b) The footprints of ICM.}
    \label{fig:flow_bonus}
\end{figure*}

\subsection{Comparison of the Model Parameters}
\label{subsec::parameter}
Table~\ref{param_table} summarizes the number of parameters used by different intrinsic curiosity modules including ICM, ICM-pixels, FICM-C, and FICM-S, as well as the results of Fig.~\ref{fig:reward}.  It shows that the catastrophic forgetting problem occurs for ICM-based method (`ICM + A3C') under both sparse reward and very sparse reward settings, while FICM-based methods (`FICM-S + A3C' and `FICM-C + A3C') show enduring performance and require fewer parameters.  Although the architectures of the flow predictors employed by FICM-S and FICM-C seem much more complicated than the forward dynamics model and inverse dynamics model in ICM, however, the number of model parameters used by FICM-S and FICM-C are relatively fewer than those used by ICM.  It can be seen that FICM-C and FICM-S require only $78.84\%$ and $55.89\%$ of the model parameters used by ICM. Please note that although the parameters of ICM-pixels are only $43.53\%$ of those used by ICM due to the removal of the inverse dynamics model, ICM-pixels fails in the environment with very sparse rewards. One explanation for the excessive number of model parameters in ICM is because of its requirement of eight input frames (four stacked frames for $S_t$ and another four for $S_{t+1}$) as its input.  On the other hand, both FICM-S and FICM-C require only a single frame for $S_t$ and another one for $S_{t+1}$, allowing their model parameters to be relatively fewer than those of ICM.



\subsection{Evolution of Intrinsic Rewards versus Time}
\label{subsec::bonus}
In order to demonstrate the effectiveness and advantage of FICM over ICM, we further conduct an experiment to demonstrate the evolution of intrinsic rewards by storing the locations that the A3C agent reached in each episode, along with the exploration bonus (i.e., intrinsic rewards).  We plot the distribution of the 3-tuples $(x, y, bonus)$ onto the game map, and illustrate the evolution of the distribution over time.  The bonus terms are reflected on the sizes and color depths of the points. In other words, a point is plotted larger and darker if its corresponding exploration bonus is high.  Ideally, in later episodes, previously visited locations should be plotted with lighter color and those newly visited should be darker.  In Fig.~\ref{fig:flow_bonus}, we can observe that FICM is capable of memorizing previously visited paths and avoiding repeated exploration.  On the other hand, even though the agent is not provided with any external reward, the baseline still explores those visited paths again in the later episodes.  More precisely, when the episode range $e$ is set to $[250,~275]$ and $[2150,~2175]$, the sudden increase in curiosity for visited paths indicates that ICM forgets previous explorations and misinterprets those visited states as novel states.  Eventually, such forgetting problem may result in the agent being unable to traverse to the goal location.  This experiment again validates that the catastrophic forgetting problem does exist in ICM, while the proposed FICM is robust to the problem.

\section{Conclusion}
\label{sec::conclusion}
We propose a flow-based curiosity module (FICM) introducing optical flow estimation, including two different implementations, FICM-C and FICM-S. FICM employs optical flow estimation errors as a suitable intrinsic reward signal for the agent to explore the environment in a more comprehensive fashion. Throughout a number of thorough experiments, FICM explores novel states efficiently and never suffers from any performance drop in either sparse or very sparse external reward settings. These experimental results demonstrate that our proposed method is capable of memorizing the novelty of states in large high-dimensional state spaces, and delivers permanent performance without encountering catastrophic forgetting problems. In one word, FICM successfully solves the forgetting problem and shows lifelong learning by well balancing exploration and exploitation via learning optical flow.


\bibliography{example_paper}
\bibliographystyle{icml2019}

\end{document}